\documentclass[11pt]{article}
\usepackage{coling2016}
\usepackage{times}
\usepackage{url}
\usepackage{latexsym}

\usepackage{comment}
\usepackage{graphicx}

\usepackage{enumitem}
\usepackage{makecell}
\usepackage{anyfontsize}
\usepackage{tabularx}
\usepackage{t1enc}
\usepackage{rotating}
\usepackage{array}
\usepackage{amsmath}
\usepackage{tablefootnote}
\usepackage{float}
\usepackage{slashbox,multirow,multicol,booktabs,pict2e}

\title{Inducing Multilingual Text Analysis Tools Using Bidirectional Recurrent Neural Networks}

\author{Othman Zennaki \\
  CEA, LIST, LVIC \\
  Gif-sur-Yvette, France \\
  othman.zennaki@cea.fr \\
  \\\And
  Nasredine Semmar \\
  CEA, LIST, LVIC \\
  Gif-sur-Yvette, France \\
  nasredine.semmar@cea.fr\\
   \\\And
  Laurent Besacier \\
  LIG, Univ. Grenoble-Alpes \\
  Grenoble, France \\
  laurent.besacier@imag.fr\\
  }

\date{}

\begin{document}
\maketitle
\begin{abstract}
This work focuses on the rapid development of linguistic annotation tools for resource-poor languages. We experiment several cross-lingual annotation projection methods using Recurrent Neural Networks (RNN) models. The distinctive feature of our approach is that our multilingual word representation requires only a parallel corpus between the source and target language. More precisely, our method has the following characteristics: (a) it does not use word alignment information, (b) it does not assume any knowledge about foreign languages, which makes it applicable to a wide range of resource-poor languages, (c) it provides truly multilingual taggers. 
We investigate both uni- and bi-directional RNN models and propose a method to include external information (for instance low level information from POS) in the RNN to train higher level taggers (for instance, super sense taggers). We demonstrate the validity and genericity of our model by using parallel corpora (obtained by manual or automatic translation). Our experiments are conducted to induce cross-lingual POS and super sense taggers.
\end{abstract}

\section{Introduction}

In order to minimize the need for annotated resources (produced through manual annotation, or by manual check of automatic annotation), several research works were interested in building Natural Language Processing (NLP) tools based on unsupervised or semi-supervised approaches \cite{Collins99,Klein05,Goldberg10}. For example, NLP tools based on cross-language projection of linguistic annotations achieved good performances in the early 2000s \cite{Yaro01}. The key idea of annotation projection can be summarized as follows: through word alignment in parallel text corpora, the annotations are transferred from the \textit{source} (resource-rich) language to the \textit{target} (under-resourced) language, and the resulting annotations are used for supervised training in the target language. However, automatic word alignment errors \cite{Fras07} limit the performance of these approaches.

Our work is built upon these previous contributions and observations. We explore the possibility of using Recurrent Neural Networks (RNN) to build multilingual NLP tools for resource-poor languages analysis. The major difference with previous works is that we do not explicitly use word alignment information. Our only assumption is that parallel sentences (source-target) are available and that the source part is annotated. In other words, we try to infer annotations in the target language from sentence-based alignments only. While most NLP researches on RNN have focused on monolingual tasks\footnote{Exceptions are the recent propositions on Neural Machine Translation \cite{Cho14,Suts14}}
 and sequence labeling \cite{Coll11,Grav12}, this paper, however, considers the problem of learning multilingual NLP tools using RNN.

\blfootnote{
    %
    %
    %
    %
     \hspace{-0.65cm}  
     This work is licensed under a Creative Commons 
     Attribution 4.0 International License.
     License details:
     \url{http://creativecommons.org/licenses/by/4.0/}
}

\textbf{Contributions} In this paper, we investigate the effectiveness of RNN architectures --- Simple RNN (SRNN) and Bidirectional RNN (BRNN) --- for multilingual sequence labeling tasks without using any word alignment information. Two NLP tasks are considered: Part Of Speech (POS) tagging and Super Sense (SST) tagging ~\cite{Ciaramita06}. Our RNN architectures demonstrate very competitive results on unsupervised training for new target languages. In addition, we show that the integration of POS information in 
RNN models is useful to build multilingual coarse-grain semantic (Super Senses) taggers. 
For this, a simple and efficient way to take into account low-level linguistic information for more complex sequence labeling RNN is proposed.
   
\textbf{Methodology} For training our multilingual RNN models, we just need as input a parallel (or multi-parallel) corpus between a resource-rich language and one or many under-resourced languages. Such a parallel corpus can be manually obtained (clean corpus) or automatically obtained (noisy corpus).

To show the potential of our approach, we investigate two sequence labeling tasks: cross-language POS tagging and multilingual Super Sense Tagging (SST). For the SST task, we measure the impact of the parallel corpus quality with  manual or automatic translations of the SemCor \cite{Miller93} translated from English into Italian (manually and automatically) and French (automatically). 

\textbf{Outline}
The remainder of the paper is organized as follows. Section \ref{Related Work} reviews related work. Section \ref{Approach} describes our cross-language  annotation projection approaches based on RNN. Section \ref{Experiments} presents the empirical study and associated results. We finally conclude the paper in Section \ref{Conclusion}.
\section{Related Work}
\label{Related Work}

Cross-lingual projection of linguistic annotations was pioneered by ~\newcite{Yaro01} who created new monolingual resources by transferring annotations from resource-rich languages onto resource-poor languages through the use of word alignments. The resulting (noisy) annotations are used in conjunction with robust learning algorithms to build cheap unsupervised NLP tools ~\cite{Pado09}. This approach has been successfully used to transfer several linguistic annotations between languages (efficient learning of POS taggers ~\cite{Das11,Duon13} and accurate projection of word senses ~\cite{Bent04}). Cross-lingual projection requires a parallel corpus and word alignment between source and target languages. Many automatic word alignment tools are available, such as GIZA++ which implements IBM models ~\cite{Och00}. However, the noisy (non perfect) outputs of these methods is a serious limitation for the annotation projection based on word alignments ~\cite{Fras07}.

To deal with this limitation, recent studies based on cross-lingual representation learning methods have been proposed to avoid using such pre-processed and noisy alignments for label projection. First, these approaches learn language-independent features, across many different languages ~\cite{Durr12,Al-R13,Tack13a,Luon15,Gouw15a,Gouw15b}. Then, the induced representation space is used to train NLP tools by exploiting labeled data from the source language and apply them in the target language. Cross-lingual representation learning approaches have achieved good results in different NLP applications such as cross-language SST and POS tagging ~\cite{Gouw15a}, cross-language named entity recognition ~\cite{Tack12}, cross-lingual document classification and lexical translation task ~\cite{Gouw15b}, cross language dependency parsing ~\cite{Durr12,Tack13a} 
and cross-language semantic role labeling ~\cite{Tito12}.  

Our approach described in next section, is inspired by these works since we also try to induce a common language-independent feature space (crosslingual words embeddings). Unlike ~\newcite{Durr12} and ~\newcite{Gouw15a}, who use bilingual lexicons, and unlike ~\newcite{Luon15} who use word alignments between the source and target languages\footnote{to train a bilingual representation regardless of the task} our common multilingual representation is very agnostic. We use a simple (multilingual) vector representation based on the occurrence of source and target words in a parallel corpus and we let the RNN learn the best internal representations (corresponding to the hidden layers) specific to the task (SST or POS tagging).
                       
In this work, we learn a cross-lingual POS tagger (multilingual POS tagger if a multilingual parallel corpus is used) based on a recurrent neural network (RNN) on the source labeled text and apply it to tag target language text. We explore simple and bidirectional RNN architectures (SRNN and BRNN respectively). Starting from the intuition that low-level linguistic information is useful to learn more complex taggers, we also introduce three new 
RNN variants to take into account external (POS) information in multilingual SST.

\section{Unsupervised Approach Overview}

\label{Approach}

To avoid projecting label information from deterministic and error-prone word alignments, we propose to represent the  word alignment information intrinsically in a recurrent neural network architecture. The idea consists in implementing a recurrent neural network as a multilingual sequence labeling tool (we investigate POS tagging and SST tagging).  
Before describing our cross-lingual (multilingual if a multi-parallel corpus is used) neural network tagger, we present the simple cross-lingual projection method, considered as our baseline in this work. 

\subsection{Baseline Cross-lingual Annotation Projection}
\label{Cross_lingual}
We use direct transfer as a baseline system which is similar to the method described in ~\cite{Yaro01}. 
First we tag the source side of the parallel corpus using the available supervised tagger. Next, we align words in the parallel corpus to find out corresponding source and target words. Tags are then projected to the (resource-poor) target language. The target language tagger is trained using any machine learning approach (we use TNT tagger ~\cite{Bran00} in our experiments).

\begin{figure*}[tb!]
\centering \includegraphics[scale=0.45]{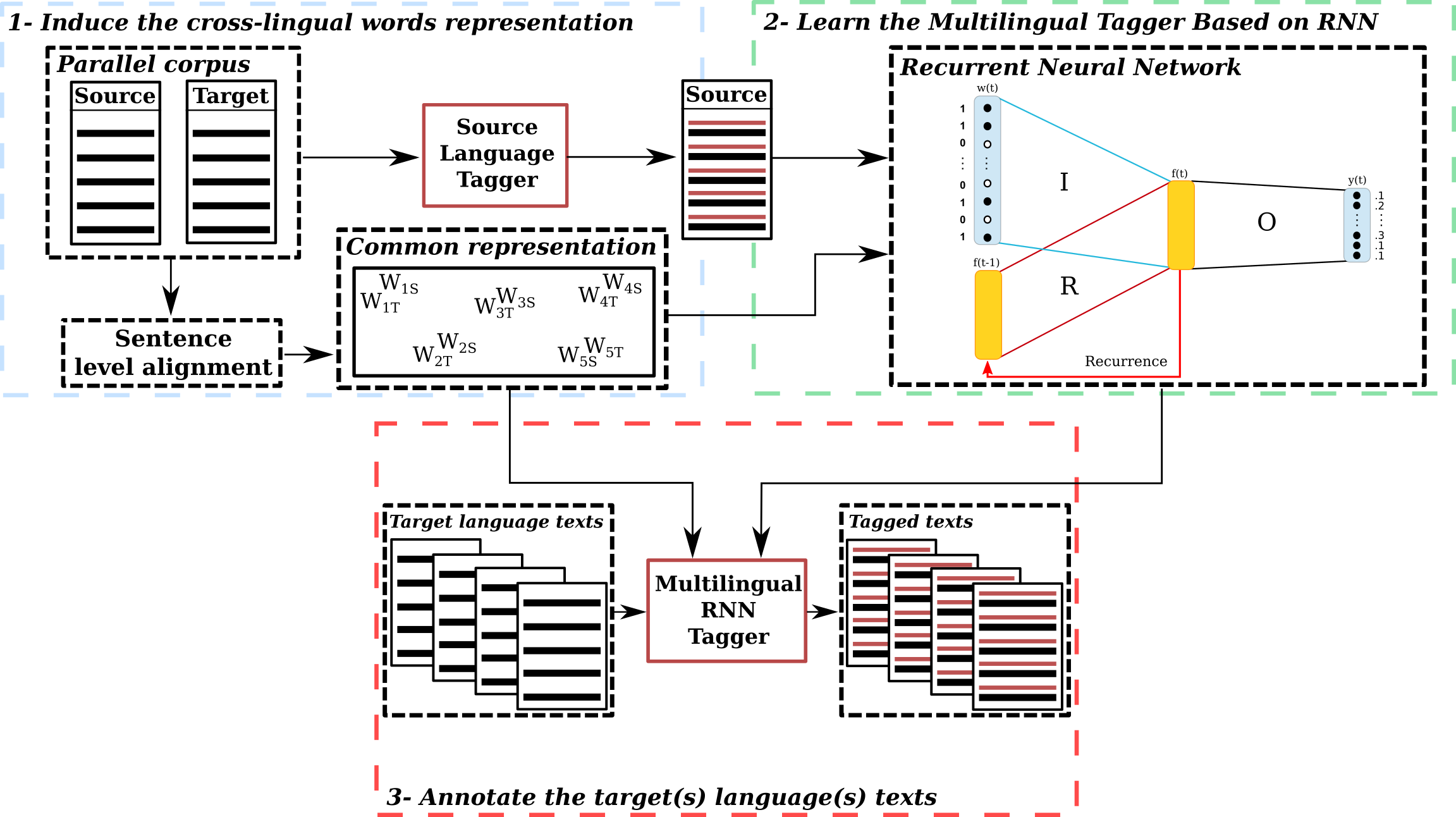}
\vspace*{-0.2cm}
\caption{\label{Method} Overview of the proposed model architecture for inducing multilingual RNN taggers.}
\vspace*{-0.3cm}
\end{figure*}

\vspace{-0.15cm}
\subsection{Proposed Approach}
\vspace{-0.15cm}
\label{Our Approach}

We propose a method for learning multilingual sequence labeling tools based on RNN, as it can be seen in Figure \ref{Method}. In our approach, a parallel or multi-parallel corpus between a resource-rich language and one or many under-resourced languages is used to extract common (multilingual) and agnostic words representations. These representations, which rely on sentence level alignment only, are used with the source side of the parallel/multi-parallel corpus to learn a neural network tagger in the source language. Since a common representation of source and target words is chosen, this neural network tagger is truly multilingual and can be also used to tag texts in target language(s).

\vspace{-0.1cm}
\subsubsection{Common Words Representation}
\vspace{-0.1cm}
In our \textit{agnostic} representation, we associate to each word (in source \textit{and} target vocabularies) a common vector representation, namely ${V_{wi}, i = 1,...,N}$, where $N$ is the number of parallel sentences (bi-sentences in the parallel corpus). If $w$ appears in i-th bi-sentence of the parallel corpus then $V_{wi}=1$. 

The idea is that, in general, a source word and its target translation appear together in the same bi-sentences and their vector representations are close. We can then use the RNN tagger, initially trained on source side, to tag the target side (because of our {\em common vector representation}). This simple representation does not require multilingual word alignments and it lets the RNN learns the optimal internal representation needed for the annotation task (for instance, the hidden layers of the RNN can be considered as multi-lingual embeddings of the words).


\subsubsection{Recurrent Neural Networks}
There are two major architectures of neural networks: Feedforward ~\cite{Beng03} and Recurrent Neural Networks (RNN) ~\cite{Schm92,Miko10}. ~\newcite{Sund13} showed that language models based on recurrent architecture achieve better performance than language models based on feedforward architecture. This is due to the fact that recurrent neural networks do not use a context of limited size. This property led us to use, in our experiments, the Elman recurrent architecture ~\cite{Elman90}, in which recurrent connections occur at the hidden layer level.

\begin{figure*}
\centering \includegraphics[scale=0.23]{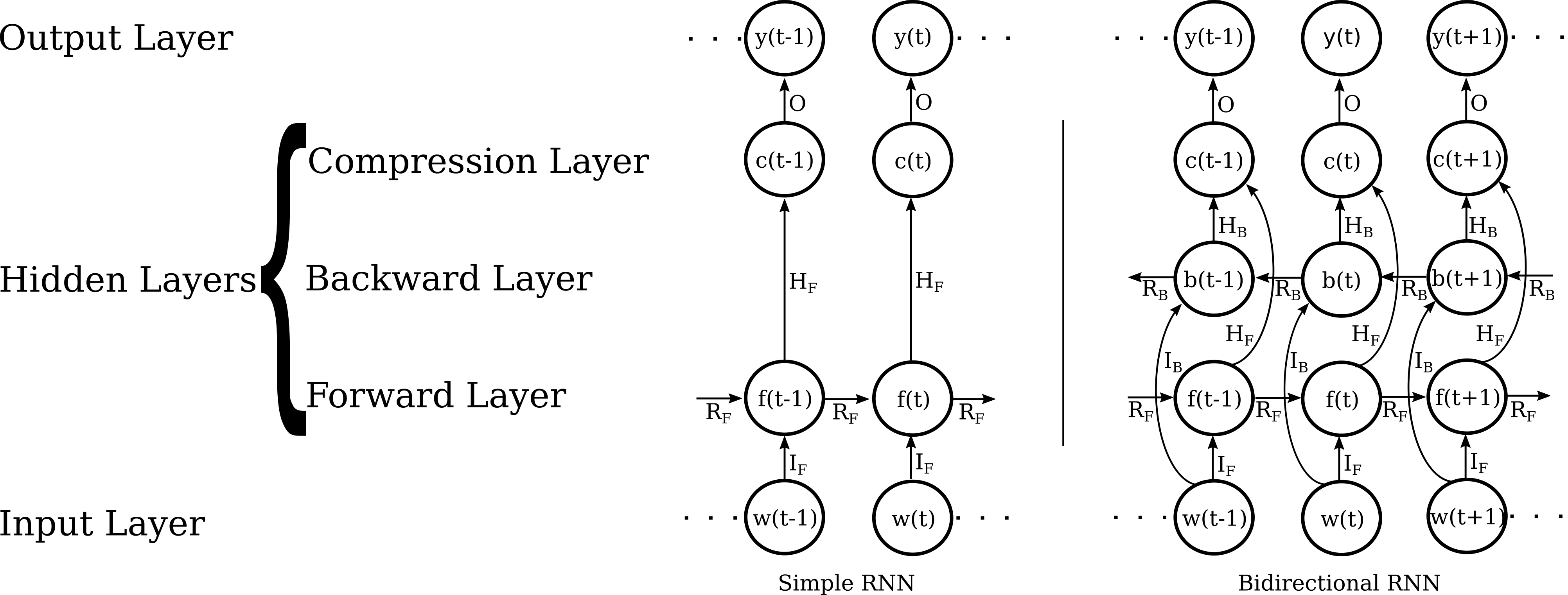}
\vspace*{-0.2cm}
\caption{\label{RNN_Architectures} High level schema of RNN used in our work.}
\vspace*{-0.2cm}
\end{figure*}

We consider in this work two Elman RNN architectures (see Figure \ref{RNN_Architectures}): \textit{Simple} RNN (SRNN) and \textit{Bidirectional} RNN (BRNN). In addition, to be able to include low-level linguistic information in our architecture designed for more complex sequence labeling tasks, we propose three new 
RNN variants to take into account external (POS) information for multilingual Super Sense Tagging (SST).

\vspace{0.2cm}\hspace{-0.4cm}\textbf{A.\hspace{0.2cm} Simple RNN} 
\vspace{0.1cm}

\hspace{-0.5cm} In the \textit{simple} Elman RNN (SRNN), the recurrent connection is a loop at the hidden layer level. This connection allows SRNN to use at the current time step hidden layer's states of previous time steps. In other words, the hidden layer of SRNN represents all previous history and not just $n-1$ previous inputs, thus the model can theoretically represent long context.

The architecture of the SRNN considered in this work is shown in Figure \ref{RNN_Architectures}. In this architecture, we have 4 layers: input layer, forward (also called recurrent or context layer), compression hidden layer and output layer. All neurons of the input layer are connected to every neuron of forward layer by weight matrix $I_F$ and $R_F$, the weight matrix $H_F$ connects all neurons of the forward layer to every neuron of compression layer    
and all neurons of the compression layer are connected to every neuron of output layer by weight matrix $O$. 

The input layer consists of a vector $w(t)$ that represents the current word $w_t$ in our common words representation (all input neurons corresponding to current word $w_t$ are set to 0 except those that correspond to bi-sentences containing $w_t$, which are set to 1), and of vector $f(t-1)$ that represents output values in the forward layer from the previous time step. We name $f(t)$ and $c(t)$ the current time step hidden layers (our preliminary experiments have shown better performance using these two hidden layers instead of one hidden layer), with variable sizes (usually 80-1024 neurons) and sigmoid activation function. These hidden layers represent our common language-independent feature space and inherently capture word alignment information. The output layer $y(t)$, given the input $w(t)$ and $f(t-1)$ is computed with the following steps : 
\begin{equation}
f(t)=\Sigma (w(t) . I_F(t) + f(t-1). R_F(t))\\
\end{equation}
\begin{equation}
c(t)=\Sigma (f(t) . H_F(t))\\
\end{equation}
\begin{equation}
y(t)=\Gamma(c(t) . O(t))\\
\end{equation}

$\Sigma$ and $\Gamma$ are the sigmoid and the softmax functions, respectively. The softmax activation function is used to normalize the values of output neurons to sum up to 1. After the network is trained, the output $y(t)$ is a vector representing a probability distribution over the set of tags. The current word $w_t$ (in input) is tagged with the most probable output tag.    

For many sequence labeling tasks, it is beneficial to have access to future in addition to the past context. So, it can be argued that our SRNN is not optimal for sequence labeling, since the network ignores future context and tries to optimize the output prediction given the previous context only. This SRNN is thus penalized compared with our baseline projection based on TNT ~\cite{Bran00} which considers both left and right contexts.
To overcome the limitations of SRNN,  a simple extension of the SRNN architecture --- namely Bidirectional recurrent neural network (BRNN) ~\cite{Schu97} --- is used to ensure that context at previous and future time steps will be considered.

\vspace{0.3cm}\hspace{-0.4cm}\textbf{B.\hspace{0.2cm} Bidirectional RNN} 
\vspace{0.2cm}

\hspace{-0.5cm} An unfolded BRNN architecture is given in Figure \ref{RNN_Architectures}. The basic idea of BRNN is to present each training sequence forwards and backwards to two separate recurrent hidden layers (forward and backward hidden layers) and then somehow merge the results. This structure provides the compression and the output layers with complete past and future context for every point in the input sequence. Note that without the backward layer, this structure simplifies to a SRNN.

\vspace{0.3cm}\hspace{-0.4cm}\textbf{C.\hspace{0.2cm} RNN Variants} 
\vspace{0.2cm}

\begin{figure} 
\begin{center}
\includegraphics[scale=0.30]{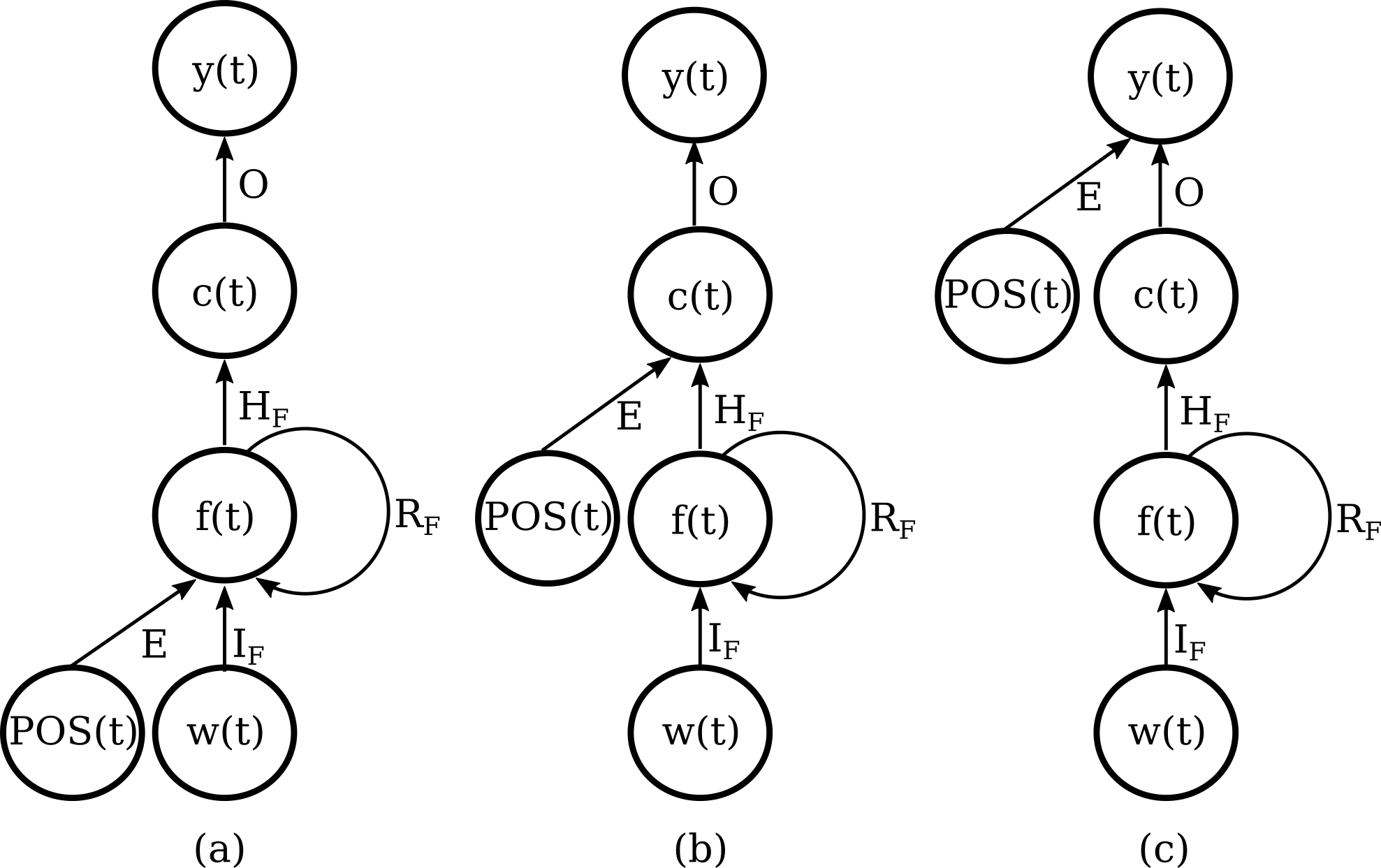}
\caption{\label{RNN_POS} SRNN variants with POS information at three levels: (a) input layer, (b) forward layer, (c) compression layer.}
\end{center}
\end{figure}

\hspace{-0.5cm} As mentioned in the introduction, we propose three new RNN variants to take into account low level (POS) information in a higher level (SST) annotation task. The question addressed here is: at which layer of the RNN this low level information should be included to improve SST performance? As specified in Figure \ref{RNN_POS}, the POS information can be introduced either at input layer or at forward layer (forward and backward layers for BRNN) or at compression layer. In all these RNN variants, the POS of the current word is also represented with a vector ($POS(t)$). Its dimension corresponds to the number of POS tags in the tagset (universal tagset of ~\newcite{Petr12} is used). We propose one \textit{hot} vector representation where only one value is set to 1 and corresponds to the index of current tag (all other values are 0).

\subsubsection{Network Training}
The first step in our approach is to train the neural network, given a parallel corpus (training corpus), and a validation corpus  (different from train data) in the source language. In typical applications, the source language is a resource-rich language (which already has an efficient tagger or manually tagged resources). Our RNN models are trained by stochastic gradient descent using usual back-propagation and back-propagation through time algorithms \cite{Rume85}. We learn our RNN models with an iterative process on the tagged source side of the parallel corpus. After each epoch (iteration) in training, validation data is used to compute per-token accuracy of the model. After that, if the per-token accuracy increases, training continues in the new epoch. Otherwise, the learning rate is halved at the start of the new epoch. Eventually, if the per-token accuracy does not increase anymore, training is stopped to prevent over-fitting. Generally, convergence takes 5--10 epochs, starting with a learning rate ${\alpha} = 0.1$.

The second step consists in using the trained model as a target language tagger (using our common vector representation).  It is important to note that if we train on a multilingual parallel corpus with \textit{N} languages ($N>2$), the same trained model will be able to tag all the \textit{N} languages. 

Hence, our approach assumes that the word order in both source and target languages are similar. In some languages such as English and  French, word order for contexts containing nouns could be reversed most of the time. For example, \textit{the European Commission} would be translated into \textit{la Commission europ\'eenne}. In order to deal with the word order constraints, 
 we also combine the RNN model with the cross-lingual projection model in our experiments.

\subsection{Dealing with out-of-vocabulary words}
For the words absent from in the initial parallel corpus, their vector representation is a vector of zero values.  Consequently, during testing, the RNN model will use only the context information to tag the OOV words found in the test corpus. To deal with these types of OOV words\footnote{words which do not have a known vector representation}, we use the CBOW model of ~\cite{Miko13distributed} to replace each OOV word by its closest known word in the current OOV word context. Once the closest word is found, its common vector representation is used (instead of the vector of zero values) at the input of the RNN.

\subsection{Combining Simple Cross-lingual Projection and RNN Models}
\label{Combined Model}
Since the simple cross-lingual projection model \textit{M1} and RNN model \textit{M2} use different strategies for tagging (TNT is based on Markov models while RNN is a neural network), we assume that these two models can be complementary. To keep the benefits of each approach, we explore how to combine them with linear  interpolation.
Formally, the probability to tag a given word \textit{w} is computed as
\begin{equation}
\small
P_{M12}(t|w)=(\mu P_{M1}(t|w,C_{M1})+(1-\mu)P_{M2}(t|w,C_{M2}))
\label{ProbCombEq}
\end{equation}
where, $C_{M1}$ and $C_{M2}$ are the context of $w$ considered by \textit{M1} and \textit{M2} respectively. The relative importance of each model is adjusted through the interpolation parameter $\mu$. The word $w$ is tagged with the most probable tag, using the function $f$ described as
\begin{equation}
f(w)= \arg\max_{t}(P_{M12}(t|w))
\label{FoncCombEq}
\end{equation}

\vspace{-0.1cm}
\section{Experiments}
\label{Experiments}
\vspace{-0.1cm}
Our models are evaluated on two labeling tasks: Cross-language Part-Of-speech (POS) tagging and Multilingual Super Sense Tagging (SST). 

\vspace{-0.1cm}
\subsection{Multilingual POS Tagging}
\vspace{-0.1cm} 
We applied our method to build RNN POS taggers for four target languages - French, German, Greek and Spanish - with English as the source language.

In order to determine the effectiveness of our common words representation described in section 3.2.1, we also investigated the use of state-of-the-art bilingual word embeddings (using MultiVec Toolkit \cite{Bera16}) as input to our RNN.

\vspace{-0.1cm}
\subsubsection{Dataset}
\vspace{-0.1cm}

For French as a target language, we used a training set of $10,000$ parallel sentences, a validation set of $1000$ English sentences, and a test set of $1000$ French sentences, all extracted from the ARCADE II English-French corpus ~\cite{vero08}. The test set is tagged with the French {\it TreeTagger} \cite{Schmid95} and  then manually checked.

For German, Greek and Spanish as a target language, we used training and validation data extracted from the Europarl corpus ~\cite{Koeh05} which are a subset of the training data used in ~\cite{Das11,Duon13}. This choice allows us to compare our results with those of  ~\cite{Das11,Duon13,Gouw15a}. The train data set contains $65,000$ bi-sentences ; a validation set of $10,000$ bi-sentences is also available. For testing, we use the same test corpora as ~\cite{Das11,Duon13,Gouw15a} (bi-sentences from CoNLL shared tasks on dependency parsing  ~\cite{Buch06}). The evaluation metric ({\it per-token} accuracy) and the ~\newcite{Petr12} \textit{universal tagset} 
are used for evaluation.

\begin{table*}[!t]
\centering
\small
	\begin{tabular}{|l||c|c||c|c|c|c|c|c|}
    \hline
    \multirow{2}{*}{\backslashbox[25mm]{\textbf{Model}}{\textbf{Lang.}}}&  \multicolumn{2}{c||}{\textbf{French}} & \multicolumn{2}{c|}{\textbf{German}} & \multicolumn{2}{c|}{\textbf{Greek}} &  \multicolumn{2}{c|}{\textbf{Spanish}} \\ \cline{2-9}

    & All words & OOV & All words & OOV & All words & OOV  & All words & OOV \\ \hline

    Simple Projection  & 80.3 & 77.1 & 78.9 &73.0  &77.5  &72.8  & 80.0  &  79.7 \\       
    \hline
    SRNN MultiVec & 75.0 & 65.4 & 70.3 & 68.8 & 71.1 & 65.4 & 73.4 & 62.4 \\       
    \hline
    \hline
    SRNN    & 78.5   & 70.0 &76.1 &76.4 & 75.7 & 70.7 & 78.8  &72.6 \\
    \hline
    BRNN  & 80.6 & 70.9 & 77.5 & 76.6 & 77.2 & 71.0 & 80.5 & 73.1 \\
    \hline
    BRNN - OOV & 81.4 & 77.8 & 77.6 & 77.8 & 77.9 & 75.3 & 80.6 & 74.7 \\
    \hline   
	Projection + SRNN &  84.5 & 78.8 & 81.5 & 77.0 &78.3 & 74.6 & 83.6 &81.2 \\
    \hline
	Projection + BRNN & 85.2 & 79.0 & 81.9 & 77.1 & 79.2 & 75.0 & 84.4 & 81.7 \\
    \hline  
    Projection + BRNN - OOV & \textbf{85.6 } & \textbf{80.4 } & 82.1 & \textbf{78.7 }  & 79.9  & \textbf{ 78.5 } & \textbf{ 84.4 } & \textbf{ 81.9 } \\
    
    \hline  
	\hline
    (Das, 2011)    & --- & --- & 82.8  & --- & \textbf{82.5 } & --- & 84.2  & ---  \\
    \hline
    (Duong, 2013)  & --- & --- & \textbf{85.4 } & --- & 80.4  & --- & 83.3 & --- \\
    \hline
    (Gouws, 2015a)  & --- & --- & 84.8  & --- & --- & --- & 82.6  & --- \\
    \hline
	\end{tabular}
\caption[The LOF caption]{\label{Tab-RNN-POS} Token-level POS tagging accuracy for Simple Projection, SRNN using MultiVec bilingual word embeddings as input, RNN\protect\footnotemark, Projection+RNN and methods of Das \& Petrov  (2011), Duong et al (2013) and Gouws \& S{\o}gaard (2015).}
\end{table*}


For training, the English (source) sides of the training corpora (ARCADE II and Europarl) and of the validation corpora are tagged with the English  {\it TreeTagger} toolkit. Using the matching provided by ~\newcite{Petr12}, we map the TreeTagger and the CoNLL tagsets to the common \textit{Universal Tagset}. 

In order to build our baseline unsupervised tagger (based on a Simple Cross-lingual Projection -- see section \ref{Cross_lingual}), we also tag the target side of the training corpus, with tags projected from English side through word-alignments established by GIZA++. After tags projection, a target language POS tagger based on TNT  approach ~\cite{Bran00} is trained. 

\footnotetext{For RNN models, only one (same) system is used to tag German, Greek and Spanish}

The combined model is built for each considered language using cross-validation on the test corpus. First, the test corpus is split into 2 equal parts and on each part, we estimate the interpolation parameter $\mu$ (Equation \ref{ProbCombEq}) which maximizes the {\em per-token} accuracy score. Then each part of test corpus is tagged using the combined model tuned 
on the other part, and vice versa (standard cross-validation procedure).

We trained MultiVec bilingual word embeddings on the parallel Europarl corpus between English and each of the target languages considered.

\subsubsection{Results and discussion}
Table \ref{Tab-RNN-POS} reports the results obtained for the unsupervised POS tagging. We note that the POS tagger based on bidirectional RNN (BRNN) has better performance than simple RNN (SRNN), which means that  both past and future contexts help select the correct tag. 
Table \ref{Tab-RNN-POS}  also shows the performance before
and after performing our procedure for handling OOVs in BRNNs. It is shown that after replacing OOVs by the closest words using CBOW, the tagging accuracy significantly increases. 

As shown in the same table, our RNN models accuracy is close to that of the simple projection tagger. It achieves comparable results to ~\newcite{Das11}, ~\newcite{Duon13} (who used the full Europarl corpus while we use only a $65,000$ subset of it) and to ~\newcite{Gouw15a} (who used extra resources such as Wiktionary and Wikipedia). Interestingly, RNN models learned using our common words representation (section 3.2.1) seem to perform significantly better than RNN models using MultiVec bilingual word embeddings. 

It is also important to note that only one single SRNN and BRNN tagger applies to German, Greek and Spanish; so this is a truly multilingual POS tagger!
Finally, as for several other NLP tasks such as language modelling or machine translation (where standard and NN-based models are generally combined in order to obtain optimal results), the combination of standard and RNN-based approaches (\textit{Projection+\_}) seems necessary to further optimize POS tagging accuracies.

\subsection{Multilingual SST}
In order to measure the impact of the parallel corpus quality on our method, we also learn our SST models using the multilingual parallel corpus MultiSemCor (MSC) which is the result of manual or automatic translation of SemCor 
from English into Italian and French.

\subsubsection{Dataset}

\textbf{SemCor} The SemCor \cite{Miller93} 
is a subset of the Brown Corpus \cite{Kucera79} labeled with the \textit{WordNet} \cite{Fellbaum98} senses.

\hspace{-0.4cm}\textbf{MultiSemCor} The English-Italian MultiSemcor (MSC-IT-1) corpus is a manual translation of the English SemCor to Italian \cite{Bent04}. As we already mentioned, we are also interested in measuring the impact of the parallel corpus quality on our method. For this we use two translation systems: (a) Google Translate to translate the English SemCor to Italian (MSC-IT-2) and French (MSC-FR-2). (b) LIG machine translation system ~\cite{Besacier12} to translate the English SemCor to French (MSC-FR-1). 
 
\hspace{-0.4cm}\textbf{Training corpus} The SemCor was labeled with the \textit{WordNet} synsets. However, because we train models for SST, we convert SemCor synsets annotations to super senses. We learn our models using the four different versions of MSC (MSC-IT-1,2 - MSC-FR-1,2), with modified Semcor on source side.

\hspace{-0.4cm}\textbf{Test Corpus}
To evaluate our models, we used the SemEval 2013 Task 12 (Multilingual Word Sense Disambiguation) \cite{Navigli13} test corpora, which are available in 5 languages (English, French, German, Spanish and Italian) and labeled with \textit{BabelNet} ~\cite{Navigli12} senses. We map BabelNet senses to WordNet synsets, then WordNet synsets are mapped to super senses.

\subsubsection{SST Systems Evaluated}
The goals of our SST experiments are twofold: first, to investigate the effectiveness of using POS information to build multilingual super sense tagger, secondly to measure the impact of the parallel corpus quality (manual or automatic translation) on our RNN models (SRNN, BRNN and our proposed variants). To summarize, we build four super sense taggers based on baseline cross-lingual projection (see section \ref{Cross_lingual}) using four versions of MultiSemcor (MSC-IT-1, MSC-IT-2, MSC-FR-1, MSC-FR-2) described above. Then we use the same four versions to train our multilingual SST models based on SRNN and BRNN. For learning our multilingual SST models based on RNN variants proposed in part (C) of section 3.2.2, we also tag SemCor using \textit{TreeTagger} (POS tagger proposed by ~\newcite{Schmid95}).

\subsubsection{Results and discussion}
Our models are evaluated on SemEval 2013 Task 12 test corpora. Results are directly comparable with those of systems which participated to this evaluation campaign. We report two SemEval 2013 (unsupervised) system results for comparison:    
\begin{itemize}
\item \textbf{MFS Semeval 2013}  :   The most frequent sense is the baseline provided by SemEval 2013 for Task 12, 
this system is a strong baseline, which is obtained by using an external resource (the WordNet most frequent sense).

\item \textbf{GETALP} : a fully unsupervised WSD system proposed by ~\cite{Schwab12ant} based on Ant-Colony algorithm. 
\end{itemize}

The DAEBAK! \cite{Navigli10} and the UMCC-DLSI systems \cite{Gutierrez11} have also participated to SemEval 2013 Task 12. However, they use a supervised approach
\footnote{DAEBAK! and UMCC-DLSI for SST have obtained: 68.1\% and 72.5\% on Italian; 59.8\% and 67.6 \% on French}.            

Table \ref{Tab-RNN-SST} shows the results obtained by our RNN models and by two SemEval 2013 WSD systems. SRNN-POS-X and BRNN-POS-X refer to our RNN variants: \textit{In} means input layer, \textit{H1} means first hidden layer and \textit{H2} means second hidden layer. 
We achieve the best performance on Italian using MSC-IT-1 clean corpus while noisy training corpus degrades SST performance. The best results are obtained with combination of simple projection and RNN which confirms (as for POS tagging) that both approaches are complementary. 

We also observe that the RNN approach seems more robust than simple projection on noisy corpora. This is probably due to the fact that no word alignments are required in our cross language RNN. Finally, BRNN-POS-H2-OOV achieves the best performance, which shows that the integration of POS information in RNN models and dealing with OOV words are useful to build efficient multilingual super senses taggers. Finally, it is worth mentioning that integrating low level (POS) information lately (last hidden layer) seems to be the best option in our case.

\begin{table*}[!t]
\centering
\small
	\begin{tabular}{|l|l||c|c||c|c|}
    \hline
    \multicolumn{2}{|c||}{\textbf{Model} } &  \multicolumn{2}{c||}{\textbf{Italian}} & \multicolumn{2}{c|}{\textbf{French}}  \\ \cline{1-6}

    \multirow{3}{*}{\begin{sideways}Baseline\end{sideways}}  &  & \textbf{MSC-IT-1} & \textbf{MSC-IT-2} & \textbf{MSC-FR-1} & \textbf{MSC-FR-2} \\ 
    & & \textbf{trans man.} & \textbf{trans. auto} & \textbf{trans. auto} & \textbf{trans auto.}\\
    \cline{2-6}
    & Simple Projection & 61.3 &  45.6 & 42.6 & 44.5 \\
    \hline
    \hline
 	\multirow{9}{*}{\begin{sideways}SST Based RNN \end{sideways}}
		& SRNN     & 59.4  &  46.2  & 46.2 & 47.0 \\
    \cline{2-6}
    & BRNN         & 59.7  &   46.2 &  46.0 &  47.2 \\
    \cline{2-6}
    & SRNN-POS-In  &  61.0   &  47.0  & 46.5  & 47.3 \\
    \cline{2-6}
    & SRNN-POS-H1  & 59.8  & 46.5  & 46.8  & 47.4 \\
    \cline{2-6}
    & SRNN-POS-H2  & 63.1  & 48.7  & 47.7 & 49.8\\
    \cline{2-6}
    & BRNN-POS-In  & 61.2  &  47.0   & 46.4  & 47.3 \\
    \cline{2-6}
    & BRNN-POS-H1  & 60.1  & 46.5  & 46.8  & 47.5 \\
    \cline{2-6}
    & BRNN-POS-H2  & 63.2  & 48.8  & 47.7  & 50 \\
    \cline{2-6}
    & BRNN-POS-H2 - OOV & 64.6  & 49.5 & 48.4  & 50.7 \\
    \hline
    \hline
    \multirow{9}{*}{\begin{sideways}Combination \end{sideways}}
    & Projection + SRNN        &  62.0          & 46.7 & 46.5  & 47.4 \\
    \cline{2-6}
    & Projection + BRNN        & 62.2         & 46.8 & 46.4  &  47.5 \\
    \cline{2-6}
    & Projection + SRNN-POS-In & 62.9         & 47.4 & 46.9  & 47.7 \\
    \cline{2-6}
    & Projection + SRNN-POS-H1 & 62.5         & 47.0   & 47.1  & 48.0 \\
    \cline{2-6}
    & Projection + SRNN-POS-H2 & 63.5         & 49.2 &  48.0   & 50.1 \\
    \cline{2-6}
    & Projection + BRNN-POS-In & 62.9         & 47.5 & 46.9  &  47.8  \\
    \cline{2-6}
    & Projection + BRNN-POS-H1 & 62.7         &  47.0  &  47.0   &   48.0 \\
    \cline{2-6}
    & Projection + BRNN-POS-H2 &63.6 & 49.3 &  48.0   &   50.3 \\
    \cline{2-6}
    & Projection + BRNN-POS-H2 - OOV &\textbf{64.7} & 49.8 &  48.6   &   51.0 \\
    \hline
	\hline
    \multirow{2}{*}{\begin{sideways}S-E\end{sideways}} & MFS Semeval 2013   & \multicolumn{2}{c||} {60.7}   & \multicolumn{2}{c|} {\textbf{52.4}} \\
    \cline{2-6}
     & GETALP \cite{Schwab12ant}   & \multicolumn{2}{c||} {40.2}  &  \multicolumn{2}{c|} {34.6}\\
    \hline

\end{tabular}
\caption{
Super Sense Tagging (SST) accuracy for Simple Projection, RNN and their combination.\label{Tab-RNN-SST}}
		\end{table*}

\section{Conclusion}
\label{Conclusion}
In this paper, we have presented an approach based on recurrent neural networks (RNN) to induce multilingual text analysis tools. We have studied Simple and Bidirectional RNN architectures on multilingual POS and SST tagging. We have also proposed new RNN variants in order to take into account low level (POS) information in a super sense tagging task. Our approach has the following advantages:  (a) it uses a language-independent word representation (based only on word co-occurrences in a parallel corpus), (b) it provides truly multilingual taggers (1 tagger for N languages) (c) it can be easily adapted to a new target language (when a small amount of supervised data is available, 
a previous study ~\cite{AnonymePACLIC2015,AnonymeTALN2015} has shown the effectiveness of our method in a weakly supervised context). 

Short term perspectives are to apply multi-task learning to build systems that simultaneously perform syntactic and semantic analysis. Adding out-of-language data to improve our RNN taggers is also possible (and interesting to experiment) with our common (multilingual) vector representation.

\bibliographystyle{acl}

\bibliography{coling2016}

\end{document}